\documentclass{article} 
\usepackage{iclr2023_conference,times}

\usepackage{amsmath,amsfonts,bm}

\def\eqref#1{equation~\ref{#1}}

\def\1{\bm{1}}

\DeclareMathAlphabet{\mathsfit}{\encodingdefault}{\sfdefault}{m}{sl}
\SetMathAlphabet{\mathsfit}{bold}{\encodingdefault}{\sfdefault}{bx}{n}

\def\gS{{\mathcal{S}}}

\def\gX{{\mathcal{X}}}

\newcommand{\E}{\mathbb{E}}

\newcommand{\R}{\mathbb{R}}

\DeclareMathOperator*{\argmax}{arg\,max}
\DeclareMathOperator*{\argmin}{arg\,min}

\usepackage{hyperref}
\usepackage{url}
\usepackage{graphicx}
\usepackage{lineno}
\usepackage[accsupp]{axessibility}  

\usepackage{tikz}
\usepackage{comment}
\usepackage{amsmath,amssymb} 
\usepackage{color}

\usepackage{graphicx}
\usepackage{amsmath,amssymb, amsthm}
\usepackage{dutchcal}
\usepackage{bbm}
\usepackage{booktabs}
\usepackage{multirow}
\usepackage{multicol}
\usepackage{xcolor}
\usepackage{enumitem}
\usepackage{wrapfig}
\usepackage{amssymb}
\usepackage{pifont}
\newcommand{\cmark}{\ding{51}}%
\setlist{nosep}
\usepackage[ruled, lined, longend, linesnumbered]{algorithm2e}
\usepackage{pifont}
\usepackage[capitalize]{cleveref}
\crefname{algorithm}{Alg.}{Algs.}
\crefname{section}{Sec.}{Secs.}
\Crefname{section}{Section}{Sections}
\Crefname{table}{Table}{Tables}
\crefname{table}{Tab.}{Tabs.}

\newcommand{\printfnsymbol}[1]{
  \textsuperscript{\star}}

\def\ourmethod{DOP\,}

\title{Fine-grained Few-shot Recognition by Deep Object Parsing}

\author{Antiquus S.~Hippocampus, Natalia Cerebro \& Amelie P. Amygdale \thanks{ Use footnote for providing further information
about author (webpage, alternative address)---\emph{not} for acknowledging
funding agencies.  Funding acknowledgements go at the end of the paper.} \\
Department of Computer Science\\
Cranberry-Lemon University\\
Pittsburgh, PA 15213, USA \\
\texttt{\{hippo,brain,jen\}@cs.cranberry-lemon.edu} \\
\And
Ji Q. Ren \& Yevgeny LeNet \\
Department of Computational Neuroscience \\
University of the Witwatersrand \\
Joburg, South Africa \\
\texttt{\{robot,net\}@wits.ac.za} \\
\AND
Coauthor \\
Affiliation \\
Address \\
\texttt{email}
}

\begin{document}

\maketitle

\begin{abstract}
We propose a new method for fine-grained few-shot recognition via deep object parsing.
In our framework, an object is made up of $K$ distinct parts and for each part, we learn a dictionary of templates, which is shared across all instances and categories. An object is parsed by estimating the locations of these $K$ parts and a set of active templates that can reconstruct the part features. 
We recognize test instances by comparing its active templates and the relative geometry of its part locations against those of the presented few-shot instances. Our method is end-to-end trainable to learn part templates on-top of a convolutional backbone. To combat visual distortions such as orientation, pose and size, we learn templates at multiple scales, and at test-time parse and match instances across these scales. We show that our method is competitive with the state-of-the-art, and by virtue of parsing enjoys interpretability as well.
\end{abstract}

\section{Introduction}
Deep neural networks (DNN) can be trained to solve visual recognition tasks with large annotated datasets. 
In contrast, training DNNs for few-shot recognition \cite{snell2017prototypical, vinyals2016matching}, and its fine-grained variant \cite{sun2020few}, where only a few examples are provided for each class by way of supervision at test-time, is challenging. 

Fundamentally, the issue is that few-shots of data is often inadequate to learn an object model among all of its myriad of variations, which do not impact an object's category. 

For our solution, we propose to draw upon two key observations from the literature.
\begin{itemize} 
\item[(A)] There are specific locations bearing distinctive patterns/signatures in the feature space of a convolution neural network (CNN), which correspond to salient visual characteristics of an image instance \cite{zhou2014object, bau2017network}.
\item[(B)] Attention on only a few specific locations in the feature space, leads to good recognition accuracy \cite{zhu2020multi, lifchitz2021few, tang2020revisiting}.
\end{itemize}

\noindent {\bf How can we leverage these observations?}\\ 
{\it Duplication of Traits.} We posit that the visual characteristics found in one instance of an object are widely duplicated among other instances, and even among those belonging to other classes. It follows from our proposition that it is the particular collection of visual characteristics arranged in a specific geometric pattern that uniquely determines an object belonging to a particular class. 

These assumptions, along with (A) and (B), imply that these shared visual traits can be found in the feature maps of CNNs and only a few locations on the feature map suffice for object recognition. CNN features are important, as they distill essential information, and suppress redundant or noisy information.

\begin{figure*}[t]
    \centering
    \includegraphics[width=0.98\linewidth]{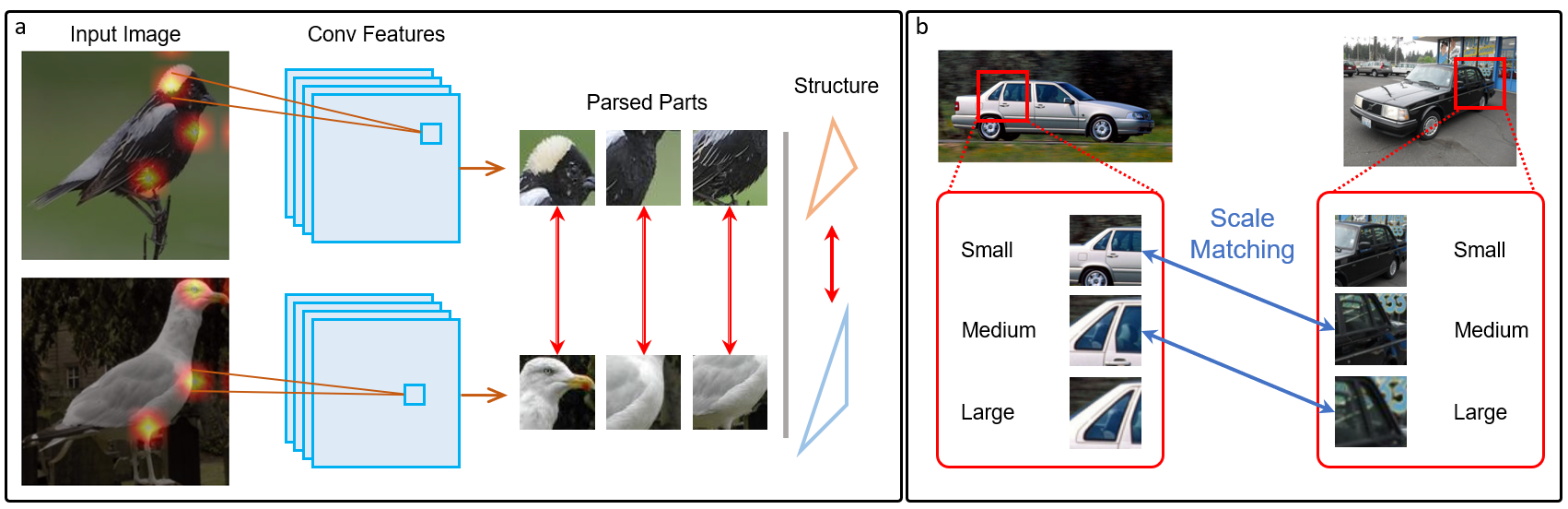}
    \caption{\textbf{Motivation:} a) In fine-grained few-shot learning, the most discriminating information is embedded in the salient parts (e.g. head and breast of a bird) and the geometry of the parts (relative part locations). Our method parses the object into a structured combination of a finite set of dictionaries, such that both finer details and the shape of the object are captured and used in recognition. b) In few shot learning, the same part may be distorted or absent in the support samples due to the perspective and pose changes. We propose to extract features and compare across multiple scales for each part to overcome this.}
    \label{fig:motivation}
    \vspace{-4mm}
\end{figure*}

\noindent {\it Parsing.} We call these finitely many latent locations on the feature maps which correspond to salient traits, {\it parts}. These parts manifest as patterns, where each pattern belongs to a finite (but potentially large) dictionary of templates. This dictionary embodies both the shared vocabulary and the diversity of patterns found across object instances. Our goal is to learn the dictionary of templates for different parts using training data, and at test-time, we seek to {\it parse}\footnote{we view our dictionary as a collection of words, parts as phrases that are a collection of words from the dictionary, and the geometric relationship between different parts as relationship between phrases.} new instances by identifying part locations and the sub-collection of templates that are expressed for the few-shot task. The provided few-shot instances are parsed and then compared against the parsed query. The best matching class is then predicted as the output. As an example see Fig \ref{fig:motivation} (a), where the recognized part locations using the learned dictionary correspond to the head, breast and the knee of the birds in their images with corresponding locations in the convolutional feature maps. In matching the images, both the constituent templates and the geometric structure of the parts are utilized.
Inferring part locations based on part-specific dictionaries is a low complexity task, and is analogous to the problem of detection of signals in noise in radar applications~\cite{van2004detection}, a problem solved by matching the received signal against a known dictionary of transmitted signals. 

\noindent {\bf Challenges.} Nevertheless, our situation is somewhat more challenging. Unlike the radar situation, we do not a-priori have a dictionary, and to learn one, we are only provided  class-level annotations by way of supervision. In addition, we require that these learnt dictionaries are compact (because we must be able to reliably parse any input), and yet sufficiently expressive to account for diversity of visual traits found in different objects and classes. 

\noindent {\it Multi-Scale Dictionaries.}
Variation in pose and orientation lead to different appearances by perspective projections, which means there is variation in the scale and size of visual characteristics of parts. 
To overcome this issue we train dictionaries at multiple scales, which leads us to a parsing scheme that parses input instances at multiple scales (see Fig. \ref{fig:motivation} (b)). 

\noindent {\it Goodness of fit.}
Besides part sizes, few-shot instances even within the same class may exhibit significant variations in poses, which can in-turn induce variations in parsed outputs. To mitigate their effects we propose a novel instance-dependent re-weighting method, for comparison, based on goodness-of-fit to the dictionary.

\noindent {\it Contributions.} We show that: (i) Our deep object parsing method,  
results in improved performance in few-shot recognition and fine-grained few-shot recognition tasks on Stanford-Car dataset outperforming prior art by  2.64\%. (ii) We provide an analysis of different components of our approach showing the effect of ablating each in final performance. Through a visualization, we show that the parts recognized by our model are salient and help recognize the object category.
\section{Related Work}
\noindent {\bf Few-Shot Classification (FSC).}
Modern FSC methods can be classified into three categories: metric-learning based, optimization-based, or data-augmentation methods. Methods in the first category focus on learning effective metrics to match query examples to support. Prototypical Network \cite{snell2017prototypical} utilizes euclidean distance on feature space for this purpose. Subsequent approaches built on this by improving the image embedding space \cite{  ye2020few,das2021importance,afrasiyabi2021mixture,zhou2021binocular,rizve2021exploring} or focusing on the metric \cite{sung2018learning, wang2019simpleshot, bateni2020improved, simon2020adaptive, li2020boosting, wertheimer2021few, fei2021z, zhang2020deepemd, zhang2021prototype}. Some recent methods have also found use of graph based methods, especially in transductive few shot classification \cite{chen2021eckpn, yang2020dpgn}. Optimization based methods train for fast adaptation using a few parameter updates with the support examples \cite{finn2017model,baik2021meta,li2017meta,lee2019meta,rajeswaran2019meta,liu2018meta}. Data-augmentation methods learn a generative model to synthesize additional training data for the novel classes to alleviate the issue of insufficient data \cite{li2020adversarial,schwartz2018delta,wang2018low,xu2021variational}.

\noindent \textbf{Fine-grained FSC.} In fine-grained few-shot classification, different classes differ only in finer visual details. An example of this is to tease apart different species of birds in images. The approaches mentioned above have been applied in this context as well \cite{li2019revisiting, sun2020few, li2020bsnet, xu2021variational}. \cite{li2019revisiting} proposes to learn a local descriptor and an image-to-class measure to capture the similarity between objects. \cite{wang2021fine} uses a foreground object extractor to exclude the noise from background and synthesize foreground features to remedy the data insufficiency. BSNet \cite{li2020bsnet} leverages a bi-similarity module to learn feature maps of diverse characteristics to improve the model's generalization ability. Variational feature disentangling (VFD) \cite{xu2021variational}, a data-augmentation method, is complementary to ours. It disentangles the feature representation into intra-class variance and class-discriminating information to emphasize the inter-class differences, and generates additional features for novel classes to mitigate data scarcity at test-time. TDM\cite{lee2022task} applies channel-wise attention to represent different classes with sparse vectors while we decompose a object into meaningful parts with vectors.

In addition, prior works on attention has been found to be effective both in few-shot learning and its fine-grained version. This approach seeks to extract most discriminative image features and ignore background information irrelevant to object class \cite{jiang2020few, zhu2020multi, lifchitz2021few, kang2021relational}. In related work, \cite{cao2020concept} propose training fixed masks on feature space, and leveraging outputs from each mask for FSC.

\noindent {\bf Recognition using Object Parts.}
Our method is closely related to recognition based on identifying object components, an approach motivated by how humans learn to recognize object \cite{biederman1987recognition}. It draws inspiration from \citet{ullman2002visual}, who showed that information maximization with respect to classes of images resulted in visual features eyes, mouth, etc. in facial images and tyres, bumper, windows, etc. in images of cars. Along these lines, Deformable Part Models (DPM) \cite{felzenszwalb2009object, felzenszwalb2010cascade} proposed to learn object models by composing part features and geometries, and utilize it for object detection. Neural Network models for DPMs were proposed in \cite{savalle2014deformable,girshick2015deformable}. Multi-attention based models, which can be viewed as implicitly incorporating parts, have been proposed~\cite{zheng2017learning} in the context of fine-grained recognition problems. Although related, a principle difference is our few-shot setting, where new classes emerge, and we need to generate new object models on-the-fly. 

Prior works on FSC have also focused on combining parts, albeit with different notions of the concept. As such, the term part is overloaded, and is unrelated to our notion. DeepEMD~\cite{zhang2020deepemd} focuses in the image-distance metric based on an earth mover's distance between different parts. Here, parts are simply different physical locations in the image and not a compact collection of salient parts for recognition. \cite{tang2020revisiting} uses salient object parts for recognition, while \cite{tokmakov2019learning} attempts to encode parts into image features. However, both these methods require additional attribute annotations for training, which may be expensive to gather and not always available. 
\cite{hao2019collect} and \cite{wu2021task} discover salient object parts and use them for recognition via attention maps similar to our method. They also re-weigh their similarity task-adaptively, which is also a feature of our method. We differ in our use of a finite dictionary of templates used to learn a compact representation of parts. Also, we use reconstruction as supervision for accurately localizing salient object parts, and impose a meaningful prior on the geometry of parts, which keeps us from degenerate solutions for part locations.

\section{Deep Object Parsing} \label{sec:sop}

\begin{figure*}[h]
    \centering
    \includegraphics[width=\linewidth]{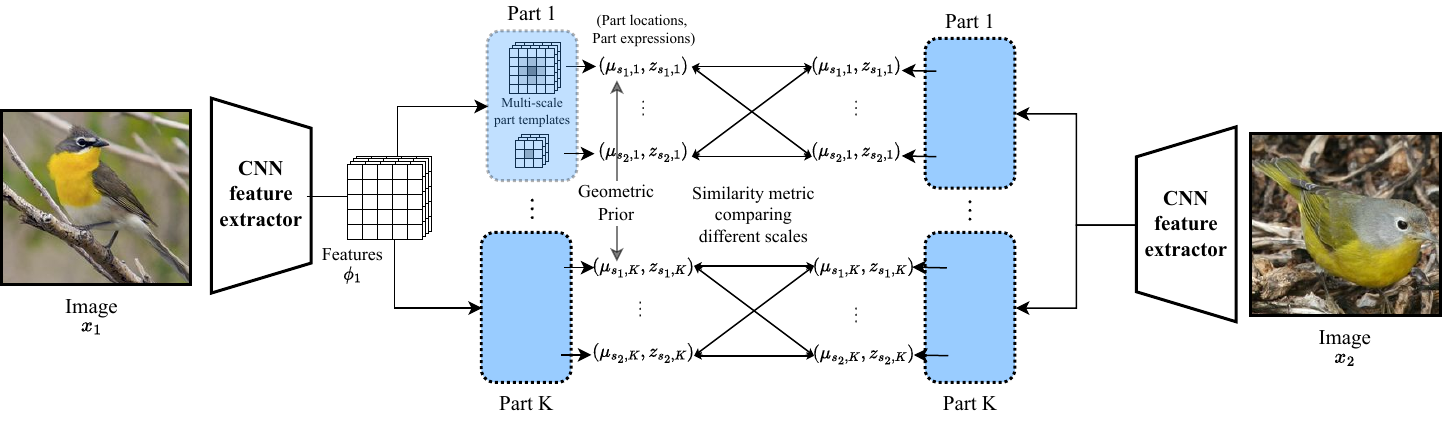}
    \caption{\textbf{Deep Object Parsing.} An image is parsed as a collection of salient parts. Each part is represented by a 2D location and part expression vector. In our method, we estimate locations and expressions at multiple scales for each part and using these, determine image similarity for few-shot recognition. (Best viewed with zoom)}
    
    \label{fig:overview}
    \vspace{-5mm}
\end{figure*}

Input instances are denoted by $x \in {\cal X}$ and we denote by $\phi \in \mathbb{R}^{G\times G \times C}$ the output features of a CNN, with $C$ channels, supported on a 2D, $G\times G$, grid.\\
\noindent {\bf Parsing Instances.}
A parsed instance has $K$ distinct parts. While this term, ``parts'', is overloaded in prior works, our notion of a part is a tuple, consisting of part-location and part-expression at that location. The location corresponds to an $s\times s$ attention mask, $M(\mu)$ centered around the location $\mu \in [G] \times [G]$ in the CNN feature space. 

Also integral to our parsing are a set of learned templates in feature space ${D}_{p,c} \in \R^{s \times s}, p \in [K], c \in [C]$. Given these templates, part-expressions are a set of reconstruction coefficients $z_{p, c}(\mu) \in \R, p \in [K], c \in [C]$ such that we reconstruct features as $\phi_{c, M_{\mu_p}} \approx z_{p, c} D_{p, c}$, where the subscript $M_{\mu_p}$ denotes a projection onto its support. We assume that the part templates are exhaustive across input categories. Once parsed, each instance is represented using its part locations $\mu_p \in [G] \times [G]$ and expressions $z_p \in \R^{C}$ for each $p \in [K]$.

\noindent 
{\bf Part Expression as LASSO Regression.}
Given an instance $x$, its feature output, $\phi$, and a candidate part-location, $\mu$, we can estimate sparse part-expression coefficients $z_p(\mu) \in \R^{C}$ by optimizing the $\ell_1$ regularized reconstruction error, at the location $\mu=\mu_p$. 
\begin{equation} \label{eq.part-exp}
z_{p}(\mu)=\arg\min_{\beta} \sum_{c \in C} \|\phi_{c, M(\mu)} - D_{p,c}\beta_{c}\|^2 + \lambda \|\beta\|_1.
\end{equation}
\noindent {\it Non-negativity.} Part expressions $z_{p,c}$ signify presence or absence of part templates in the observed feature vectors, and as such can be expected to take on non-negative values. This fact turns out to be useful later for DNN implementation.

\noindent {\bf Part Location Estimation.} Note that part expression $z_{p}$ is a function of location $\mu$, while the part location $\mu_p$ can be estimated by plugging in the optimal part-expressions for each candidate location value, namely, 
\begin{equation} \label{eq.part-location}
\mu_p = \argmin_{\mu \in [G] \times [G]}\sum_{c \in C} \|[\phi_c]_{M(\mu)} - D_{p,c}z_{p,c}(\mu)\|^2 + \lambda \|z_{p}(\mu)\|_1
\end{equation}
This couples the two estimation problems, and is difficult to implement with DNNs, motivating our approach below.

\noindent {\bf Feedforward DNNs for Parsing.}
To make the proposed approach amenable to DNN implementation, we approximate the solution to Eq.~\ref{eq.part-exp} by optimizing the reconstruction error followed by thresholding, namely, we compute 
$z_p^\prime (\mu)=\arg\min_{\beta} \sum_{c \in C} \|\phi_{c, M(\mu)} - D_{p,c}\beta_{c}\|^2$, and we threshold the resulting output by deleting entries smaller than $\zeta$: $S_{\zeta}(u)=u\mathbf{1}_{|u|\geq \zeta}$. This is closely related to thresholding methods employed in LASSO~\cite{hastie01statisticallearning}.

The quadratic component of the loss allows for an explicit solution, and the solution reduces to template matching per channel, which can further be expressed as a convolution~\cite{gonzalez2008digital}. Using this insight, we derive our estimate of $\mu_p$ as
\begin{align}
    \mu_p = \argmax_{\mu \in [G] \times [G]} \sum\limits_{c\in C}((\theta_{p,c}\ast\phi_c)(\mu)-\lambda_c)^2  \label{eq.estimate_mu0}
\end{align}
where $\theta_{p,c}=D_{p,c}/\|D_{p,c}\|$, and $\lambda_c = \lambda/2\|D_{p,c}\|$ becomes a channel dependent constant. With the above estimate of $\mu_p$, we get the estimate of $z_p$ as:
\begin{equation}
z_{p,c}^\prime = \frac{(D_{p,c} \ast \phi_c)(\mu_p)}{\|D_{p,c}\|^2}; \,\,\, z_{p,c}(\mu)=S_{\zeta}(z_{p,c}^\prime) \label{eq.estimate_z0}
\end{equation}
where `$:$' is the double dot product---sum over all entries of the element-wise (Hadamard) product and $z_{p,c}$ is the $c^{th}$ element of $z_p$.
For a full derivation of the above estimates, please refer to the supplementary materials (App. A).

\noindent {\it Estimates differentiable in parameters.} Since argmax is a non-differentiable function, using \cref{eq.estimate_mu0} for estimating part-locations does not allow us to use gradient based learning for the parameters of the DNN. We can circumvent this by approximating the argmax as the expectation of a softmax distribution $\nu_p$ over $[G] \times [G]$ with a low temperature $T$. 
\begin{align}
\nu_p(\mu) = \text{softmax} \left( \frac{1}{T} \sum\limits_{c\in C}((\theta_{p,c}\ast\phi_c)(\mu)-\lambda_c)^2\right); \quad \mu_p = \E_{\mu \sim \nu_p} \mu
\label{eq.estimate_mu}
\end{align}
Similarly, we get a differentiable estimate of part expressions as
\begin{align}
    z_{p,c}^\prime = \E_{\mu \sim \nu_p} \left[\frac{(D_{p,c} \ast \phi_c)(\mu)}{\|D_{p,c}\|^2} \right]; \,\,\, z_{p,c}=S_{\zeta}(z_{p,c}^\prime) \label{eq.estimate_z}
\end{align}

\noindent {\it Multi-Scale Extension.} We extend our approach to incorporate parsing parts at multiple scales. This is often required because of significant difference in orientation and pose between query and support examples. To do so we simply consider masks $M(\mu)$ and templates $D$ at varying mask sizes $s \in \gS$, each leading to independent part location and expression estimates ($\mu_{s, p}$, $z_{s, p}$) for part $p$. \cref{alg:alg1} specifies the parse of an input instance and \cref{fig:overview} shows an overview of object parsing.

\begin{algorithm}[h]
\DontPrintSemicolon
 \textbf{Given:} Convolutional backbone $f$, templates $\{D_{s,p,c}\}_{s\in\gS,p\in[K],c\in[C]}$, threshold $\zeta$, $\ell_1$ regularization constant $\lambda$, temperature $T$  \\
 \textbf{Input:} Image $x$ \\
 Compute convolutional features $\phi = f(x)$ \\
 \For{$p \in [K], s \in \gS$}{
    Estimate $\mu_{s, p}$ (through $\nu_{s,p}$) using \cref{eq.estimate_mu} \\
    Estimate $z_{s, p} = [z_{s,p,c}]_{c\in[C]}$ using \cref{eq.estimate_z} 
 }
 
 \textbf{Output}: Part locations and expressions ($\{\mu_{s,p}\}_{p\in[K], s\in \gS}$, $\{z_{s,p}\}_{p\in[K],s\in\gS}$)
 \caption{\texttt{PARSE} (Object Parsing using DNNs)}
 \label{alg:alg1}
\end{algorithm}

\subsection{Few-Shot Recognition} 

At test-time we are given a query instance, $q$, and by way of supervision, $M$ support examples each for $N$ classes, and the goal is to predict the query class label $y^{(q)} \in [N]$. We first run \texttt{PARSE} (\cref{alg:alg1}) on each of these. $\texttt{PARSE}(q) = (\{\mu_{s,p}^{(q)}\}, \{z_{s,p}^{(q)})\}$ and for the $i^{th}$ support example of class $y$, $\texttt{PARSE}(x^{(i, y)}) = (\{\mu_{s,p}^{(i,y)}\}, \{z_{s,p}^{(i,y)}\})$. For comparing query and support examples we need a notion of distance/similarity, which we define next. 

\noindent{\it Goodness-of-fit reweighting.} The entropy of the distribution $\nu_{s,p}$ is an important indicator of goodness-of-fit of the dictionary templates (lower entropy meaning a more precise and confident part-location prediction as a result of a better fit). Let $h_{s,p}^{(q)}$ and $h_{s,p}^{(i, y)}$ be the entropies of $\nu_{s,p}^{(q)}$ and $\nu_{s,p}^{(i, y)}$ respectively. To use these as weights for computing distance (as below), we learn a simple parametric function $\alpha : \R^{M+1} \rightarrow \R$. 

Additionally, with $z_{s,p}^{(y)} =  \sum_{i\in[M]} z_{s,p}^{(i, y)}, s\in\gS, p\in[K]$ we represent the mean part expression over all support examples in class $y$. With these, we define the following distance measure between the query example $q$ and the support examples of class $y$.
\begin{align} \label{eq.dist}
    d(q,y) =& \sum_{p \in [K]} \sum_{s_1 \in \gS} \sum_{s_2 \in \gS} \alpha(h_{s_1, p}^{(q)},[h_{s_2,p}^{(i,y)}]_{i\in[M]})\left \| z_{s_1, p}^{(y)}-z_{s_2, p}^{(q)} \right \|^2  \quad \text{[expression distance]} \nonumber\\
    &+ \gamma \sum_{i\in[M]} \sum_{s_1 \in \gS} \sum_{s_2 \in \gS} \left \|\psi([\mu_{s_1, p}^{(i, y)}]_{p\in[K]})-\psi([\mu_{s_2,p}^{(q)}]_{p\in[K]})\right \|^2  \quad \text{[geometric distance]}
\end{align}
where $\psi([\mu_{s,p}]_{p\in[K]})$ is a vector of pairwise distances between all part locations at scale $s$, normalized to unit sum. The distance function consists of an expression term, and a geometric term with $\gamma$ acting as a tunable weight to control the proportion of the two. Each term is a sum over all combinations of part scales over query and support. Note that the geometric term simply attempts to find if two polygons with vertices at part locations are similar (i.e. have proportional sides), with the distance being $0$ if they are. Finally, the class prediction is made as $\widehat{y}^{(q)} = \argmin_{y \in [N]} d(q, y)$.

\noindent{\bf Training.} We train in episodes following convention. For each episode, we sample $N$ classes at random, and additionally sample support and query examples belonging to these classes from training data (details in \cref{sec:experiments}). Using a softmax over the negative distance function above as the class distribution of query $q$, we define the cross-entropy loss as
\begin{align} \label{eq.ce}
    \ell_{CE}(q) = -\log \frac{\exp(-d(q, y^{(q)}))}{\sum_{y \in [N]} \exp(-d(q, y))}
\end{align}
Additionally, while training, we impose a geometric prior to get diverse instance parts in \texttt{PARSE} by maximizing the Hellinger distance~\cite{cup} $\mathbb{H}(\cdot, \cdot)$ between part distributions. The corresponding criterion for minimization is 
\begin{align} \label{eq.div}
    \ell_{div}(x) = - \sum_{s \in \gS} \sum_{\substack{p, p^\prime \in [K] \\ p \neq p^\prime}} \mathbb{H}(\nu_{s, p}, \nu_{s, p^\prime})
\end{align}
\cref{alg:alg2} outlines the loss computation for a single query $q$. In \cref{alg:loss}, $\eta$ is a tunable parameter controlling the weight of the prior. In each episode, we use an average of the loss output over multiple query examples, which results in an end-to-end differentiable criterion in all trainable parameters, allowing us to optimize using gradient descent.

\begin{algorithm}[h]
\DontPrintSemicolon
 \textbf{Given:} requirements for \cref{alg:alg1} \texttt{PARSE}, weighting function $\alpha$, tunable parameter $\gamma$ \\
 \textbf{Input:} Query example with ground truth label $q, y^{(q)} \in \gX \times [N]$. Support examples $I = \bigcup_{y\in[N]} \left[I_y = \{x^{(i,y)}\}_{i\in[M]}\right]$ \\
 \textbf{Trainable parameters:} Convolutional backbone $f$, part templates $\{D_{s,p,c}\}_{s\in\gS,p\in[K],c\in[C]}$, weighting function $\alpha$ \\

 Compute parses $\texttt{PARSE}(q) = (\{\mu_{s,p}^{(q)}\}, \{z_{s,p}^{(q)})\}$, $\texttt{PARSE}(x^{(i, y)}) = (\{\mu_{s,p}^{(i,y)}\}, \{z_{s,p}^{(i,y)}\}) ; s\in\gS, p\in[K]$ \\
 Compute distances $d(q, y)$ for $y \in [N]$ using \cref{eq.dist} \\
 \textbf{Output}: loss $\ell(q) = \ell_{CE}(q) + \eta \frac{1}{|I|+1} \sum_{x \in I \cup \{q\}}\ell_{div}(x)$ using \cref{eq.ce,,eq.div} 
 \label{alg:loss}
 \caption{Training loss for Few-shot recognition with DOP (single episode, single query)}

 \label{alg:alg2}
\end{algorithm}

\vspace{-0.1in}
\subsection{Implementation Details}
We use two Resnet~\cite{he2016deep} backbones (Resnet-12 and Resnet-18) as feature extractors. We use the Resnet-12 model used in prior works \cite{lee2019meta}, which has more output channels and more parameters (12.4M) compared to Resnet-18 (11.2M). The input image is resized to 84 $\times$ 84 for Resnet-12 and 224 $\times$ 224 for Resnet-18. In the output features, for Resnet-12, $C = 640$ and $G = 5$, and for Resnet-18, $C = 512$ and $G = 7$. 

The number of parts $K$ is set to 4 for most experiments (see \cref{sec:ablation} for a experiments with different number of parts).

There are three scales $\gS = \{1, 3, 5\}$ considered for each part. The temperature $T$ in \cref{eq.estimate_mu} is set to $0.01$ and the threshold $\zeta$ in \cref{eq.estimate_z} is set to $0.05$. $\gamma$ is set to 0.01. For $\alpha$ we used a linear layer, and subsequently normalized $[\alpha(h_{s_1, p}^{(q)},[h_{s_2,p}^{(i,y)}]_{i\in[M]})]_{p\in[K]}$(a K-dimensional vector; see \cref{eq.dist}) with a softmax.

\section{Experiments}
\label{sec:experiments}

\subsection{Fine-grained Few-Shot Classification}
\noindent We compare \ourmethod on four fine-grained datasets: Caltech-UCSD-Birds (CUB) \cite{WahCUB_200_2011}, Stanford-Dog (Dog) \cite{stanford_dog} Stanford-Car (Car) \cite{stanford_car} and Aircraft \cite{maji2013fine} against state-of-the-art methods.

\noindent {\bf Caltech-UCSD-Birds (CUB)} \cite{WahCUB_200_2011} is a fine-grained classification dataset with 11,788 images of 200 bird species. Following convention\cite{hilliard2018few}, the 200 classes are randomly split into 100 base, 50 validation and 50 test classes. 

\noindent {\bf Aircraft} contains 100 classes of aircrafts and 10,000 images in total. Following recent benchmark \cite{lee2022task} \cite{wertheimer2021few}, we processes all images based on bounding box. And the 100 classes are split into 50, 25 and 25 classes for training, validation and test.

\noindent {\bf Stanford-Dog/Car} \cite{stanford_dog,stanford_car} are two datasets for fine-grained classification. Dog contains 120 dog breeds with a total number of 20,580 images, while Car consists of 16,185 images from 196 different car models. For few-shot learning evaluation, we follow the benchmark protocol proposed in \cite{li2019revisiting}. Specifically, 120 classes of Dog are split into 70, 20, and 30 classes, for training, validation, and test, respectively. Similarly, Car is split into 130 train, 17 validation and 49 test classes.

\noindent \textbf{Experiments Setup.}
\label{sec:exp_setup}
We conducted 5-way (5 classes episode) 1-shot and 5-way 5-shot classification tasks on all datasets. Following the episodic evaluation protocol in \cite{vinyals2016matching}, at test time, we sample 600 episodes and report the averaged Top-1 accuracy. In each episode, 5 classes from the test set are randomly selected. 1 or 5 samples for each class are sampled as support data, and another 15 examples are sampled for each class as the query data. The model is trained on train split and the validation split is used to select the hyper-parameters. We compare our \ourmethod to state-of-the-art few-shot learning methods~\cite{kang2021relational, wertheimer2021few, lee2022task, zhang2020deepemd} and fine-grained few shot learning methods: FOT\cite{wang2021fine}, VFD \cite{xu2021variational}, DN4\cite{li2019revisiting} and TDM\cite{lee2022task}. More details on these methods are in the supplementary (App. B).

\noindent \textbf{Training Details.}
Our model is trained with 10,000 episodes on CUB and 30,000 episodes on Stanford-Dog/Car for experiments with both ResNet12 and ResNet18. In each episode, we randomly select 10 classes and sample 5 and 10 samples as support and query data. The weight on the geometric prior $\eta$ is set to 1.0 on CUB and 0.1 on Stanford-Dog/Car, respectively. 

We train from scratch with Adam optimizer \cite{kingma2014adam}. The learning rate starts from 5e-4 on CUB and 1e-3 on Stanford-Car/Dog, and decays to 0.1x every 3,000 episodes on CUB and 9,000 episodes on Dog/Car. On CUB, objects are cropped using the annotated bounding box before resizing to the input size. On Stanford-Car/Dog, we use the resized raw image as the input. We employed standard data augmentations, including horizontal flip and perspective distortion, to the input images.

\begin{table}[ht!]
    \centering
    \caption{Few-shot accuracy in $\%$ on CUB (along with 95\% confidence intervals). If not specified, the results is reported by the original paper. *: results reported in \cite{xu2021variational}. $\dagger$: results are obtained by running the public implementations released by authors using ResNet18 backbone.}
    \vspace{1mm}
    \renewcommand{\arraystretch}{1.1}
    \setlength{\tabcolsep}{2mm}
    \begin{tabular}{l c c c}
    \toprule
    \bf Methods & \bf Backbone & \it 1-shot & \it 5-shot \\
    \midrule
    ProtoNet\cite{snell2017prototypical} &ResNet18 & 71.88$\pm$0.91 & 87.42$\pm$0.48 \\
    Baseline++\cite{chen2019closer} &ResNet18 & 67.02$\pm$0.9 & 83.58$\pm$0.54 \\
    SimpleShot\cite{wang2019simpleshot} &ResNet18 & 62.85$\pm$0.20 & 84.01$\pm$0.14 \\
    DN4\cite{li2019revisiting}$\dagger$ &ResNet18 & 70.47$\pm$0.72 & 84.43$\pm$0.45 \\
    AFHN\cite{li2020adversarial} &ResNet18 & 70.53$\pm$1.01 & 83.95$\pm$0.63 \\
    $\Delta$-encoder\cite{schwartz2018delta} &ResNet18 & 69.80$\pm$0.46 & 82.60$\pm$0.35 \\
    BSNet\cite{li2020bsnet} & ResNet18 & 69.61$\pm$0.92 & 83.24$\pm$0.60 \\
    FOT\cite{wang2021fine} & ResNet18 & 72.56$\pm$0.77 & 87.22$\pm$0.46 \\
    MTL\cite{liu2018meta}* &ResNet12 & 73.31 $\pm$ 0.92 & 82.29 $\pm$0.51\\
    MetaOptNet\cite{lee2019meta}* &ResNet12 & 75.15$\pm$0.46 & 87.09$\pm$0.30 \\
    DeepEMD\cite{zhang2020deepemd} &ResNet12 & 75.65$\pm$0.83 & 88.69$\pm$0.50 \\
    VFD \cite{xu2021variational} &ResNet12 & 79.12$\pm$0.83 & 91.48$\pm$0.39 \\
    FRN\cite{wertheimer2021few} & ResNet12 & 83.16 & 92.59 \\
    RENet\cite{kang2021relational} &ResNet12 & 79.49$\pm$0.44 & 91.11$\pm$0.24 \\
    TDM\cite{lee2022task} &ResNet12 & 83.36 & 92.80\\
    \midrule
    \ourmethod &ResNet18 &82.62$\pm$0.65 & \bf{92.61$\pm$0.38} \\
    \ourmethod & ResNet12 & \bf 83.39$\pm$0.82 & \bf 93.01$\pm$0.43 \\
    \bottomrule
    \end{tabular}
    \label{tab:result_cub}
    \vspace{-4mm}
\end{table}

\begin{wraptable}[13]{r}{7cm}

\vspace{-2mm}
    \caption{Few-shot classification accuracy in \% on Aircraft dataset. Unless specified, methods use ResNet-12 backbone.}
    \vspace{2mm}
    \scalebox{0.85}{
    \centering
    \renewcommand{\arraystretch}{1.1}
    \setlength{\tabcolsep}{2mm}
    \begin{tabular}{l c c }
    \toprule

    {\bf Methods} & \it 1-shot & \it 5-shot  \\
    \midrule
    ProtoNet\cite{snell2017prototypical}&67.28 &83.21\\
    DSN\cite{simon2020adaptive}& 70.23 & 83.03\\
    CTX\cite{doersch2020crosstransformers}& 71.57 & 79.31 \\
    FRN\cite{wertheimer2021few} & 69.58 & 83.98\\
    TDM\cite{lee2022task} & 71.57 & 84.77 \\

    \midrule
    \ourmethod (ResNet18)& \bf84.26  & \bf 93.41 \\
    \ourmethod (ResNet12) & \bf 85.50  & \bf 94.35\\
    \bottomrule
    \end{tabular}
    }
    \label{tab:result_air}
    \vspace{-2mm}
\end{wraptable}

\noindent \textbf{Results.}
Our results along with comparisons against state-of-the-art on CUB, Aircraft and Stanford-Dog/Car are reported in \cref{tab:result_cub,,tab:result_air,,tab:result_car}, respectively. 

\noindent {\bf \ourmethod{} is competitive with or outperforms recent works on fine-grained FSC.}
On CUB (\cref{tab:result_cub}), \ourmethod outperforms all compared approaches with 83.39\% 1-shot accuracy and 93.01\% for 5-shot with Resnet-12 backbone. Same is the case for Aircraft (\cref{tab:result_air}), where DOP outperforms a recent method in TDM~\cite{lee2022task} by a large margin using ResNet-12 backbone, performing at 85.50\% 1-shot and 94.35\% 5-shot accuracy. While on Stanford-Car (\cref{tab:result_car}), we outperform compared approaches by 3.06\% and 1.85\% on 1-shot and 5-shot, respectively, on Stanford-Dog, we outperform all methods but VFD. We note here that VFD generates additional features at test-time for novel classes, and is as such, complementary to \ourmethod. 

\vspace{-2mm}
\begin{table*}[h]
    \caption{Few-shot classification accuracy in \% on Stanford-Car/Dog benchmarks (along with 95\% confidence intervals). *: results reported in \cite{xu2021variational}. $\dagger$: results are obtained by running the codes released by authors using ResNet18 backbone.}
    \vspace{1mm}
    \scalebox{0.85}{
    \centering
    \renewcommand{\arraystretch}{1.1}
    \setlength{\tabcolsep}{2mm}
    \begin{tabular}{l c c c c c}
    \toprule
    \multirow{2}{*}{\bf Methods} &\multirow{2}{*}{\bf Backbones}& \multicolumn{2}{c}{\bf Car} & \multicolumn{2}{c}{\bf Dog}\\
    & & \it 1-shot & \it 5-shot & \it 1-shot & \it 5-shot \\
    \midrule
    ProtoNet$\dagger$\cite{snell2017prototypical}&ResNet18 &60.67$\pm$0.87 &75.56$\pm$0.45 &61.06$\pm$0.67 &74.31$\pm$0.51 \\
    DN4$\dagger$\cite{li2019revisiting}&ResNet18 & 78.77$\pm$0.81 & 91.99$\pm$0.41 & 60.73$\pm$0.67 & 75.33$\pm$0.38 \\
    MetaOptNet$\dagger$\cite{lee2019meta}& ResNet18 & {60.56}$\pm$0.78 & {76.35}$\pm$0.52 & 65.48$\pm$0.56 & 79.39$\pm$0.43 \\
    MTL*\cite{liu2018meta}&ResNet12 & - & - & 54.96$\pm$1.03 & 68.76$\pm$0.65 \\
    $\Delta$-encoder*\cite{schwartz2018delta}&ResNet12 & - & - & 68.59$\pm$0.53 & 78.60$\pm$0.78 \\
    BSNet\cite{li2020bsnet}&ResNet12 & 60.36$\pm$0.98 & 85.28$\pm$0.64 & 69.09$\pm$0.90 & 82.45$\pm$0.58 \\
    VFD*\cite{xu2021variational}&ResNet12 & - & - & \bf 76.24$\pm$0.87 & \bf 88.00$\pm$0.47 \\
    
    \midrule
    \ourmethod &ResNet18& \bf 81.41$\pm$ 0.71 & \bf 93.48$\pm$ 0.38 & 70.56$\pm$ 0.75 & 84.75$\pm$ 0.41 \\
    \ourmethod & ResNet12 & \bf 81.83$\pm$ 0.78 & \bf 93.84$\pm$ 0.45 & 70.10$\pm$ 0.79 & 85.12$\pm$ 0.55 \\
    \bottomrule
    \end{tabular}
    }
     
    \label{tab:result_car}
\end{table*}

\subsection{Analyses}
\label{sec:ablation}
\noindent We conducted a series of analyses to study the effects of different components of \ourmethod{} on fine-grained datasets based on 5-shot accuracy with the Resnet-18 backbone.

\noindent {\bf Instance-dependent reweighting based on goodness-of-fit.} We use a parametric reweighting function $\alpha$ that reweights the distances between part expressions based on the how well the learned templates fit the part features (see \cref{eq.dist}). In \cref{tab:ablation_reweighting}, we show the effect of removing this reweighting, and simply using an average of all pairs of distances between the query and support. As we see the reweighting function does help FSL accuracy. 

\noindent {\bf Effect of using part-geometry for comparison.} In \cref{eq.dist}, we use part geometries besides part expressions for computing distances. \cref{tab:ablation_reweighting} also shows scenarios where we remove this component in the distance (equivalent to setting $\gamma=0$). We see that using a distance between part geometries help final FSL performance.

\begin{table}[h]
\centering
\caption{5-way 5-shot accuracy on ablating components in distance computation: re-weighting function $\alpha$ and using part-geometry (\cref{eq.dist}).  Both components help FSL accuracy independently as well as together.}
\vspace{1mm}
\begin{tabular}{c c|c c c}
        \toprule
        \bf Part-geometry & \bf Re-weighting & \bf CUB & \bf Dog &\bf Car \\
        \midrule
         & & 91.83 & 82.07 & 92.78\\ 
          & \cmark & 92.44 & 83.90 & 93.31\\ 
        \cmark & & 91.95 &83.33 &93.21\\
        \midrule
        \cmark & \cmark & \bf 92.61 & \bf 84.75 & \bf 93.48\\
        \bottomrule
    \end{tabular}
    \label{tab:ablation_reweighting}
    \vspace{-2mm}
\end{table}

\noindent {\bf Effect of using templates at multiple scales.} In \cref{tab:ablation_scales} we ablated different choices of scales on Dog. Using multiple scales is better than a single scale, and using all the scales obtains the best performance. This validates our hypothesis that parts are distorted due to pose variations, and a single scale is not sufficient to represent the object in a few-shot scenario.

\noindent {\bf Effect of using different number of parts.} \cref{tab:ablation_parts} shows the effect of using different number of parts on 5-way 5-shot accuracy on CUB. $K=4$ has the best accuracy and performance drops with more parts as the model starts learning irrelevant or background signatures.

\begin{table}[h]
    \parbox{.48\linewidth}{
    \centering
    \caption{Effect of using templates at different scales on 5-way 5-shot classification accuracy on Dog.}
    \vspace{1mm}
    \renewcommand{\arraystretch}{1.1}
    \setlength{\tabcolsep}{2mm}
    \begin{tabular}{l|c c c c}
        \toprule
        \bf Scales & [3] & [5] & [3,5] & [1,3,5] \\
        \midrule
        \bf Accuracy & 81.56 &81.38 & 83.04 & \bf 84.75 \\
        \bottomrule
    \end{tabular}
    \label{tab:ablation_scales}
    }
    \hfill
    \parbox{.48\linewidth}{
    \centering
    \caption{Effect of using different number of parts on 5-way 5-shot classification accuracy on CUB.}
    \vspace{1mm}
    \renewcommand{\arraystretch}{1.1}
    \setlength{\tabcolsep}{1.5mm}
    \begin{tabular}{l|c c c c c}
        \toprule
        \bf Num parts & 3 & 4 & 5 & 6 \\
        \midrule
        \bf Accuracy & 92.10 & \bf 92.61 & 92.21 & 92.06 \\ 
        \bottomrule
    \end{tabular}
    \label{tab:ablation_parts}
    }
    \vspace{-2mm}
\end{table}

\begin{figure}[h]
    \centering
    \includegraphics[trim=0 4.7cm 0 2.2cm,clip,width=1\linewidth]{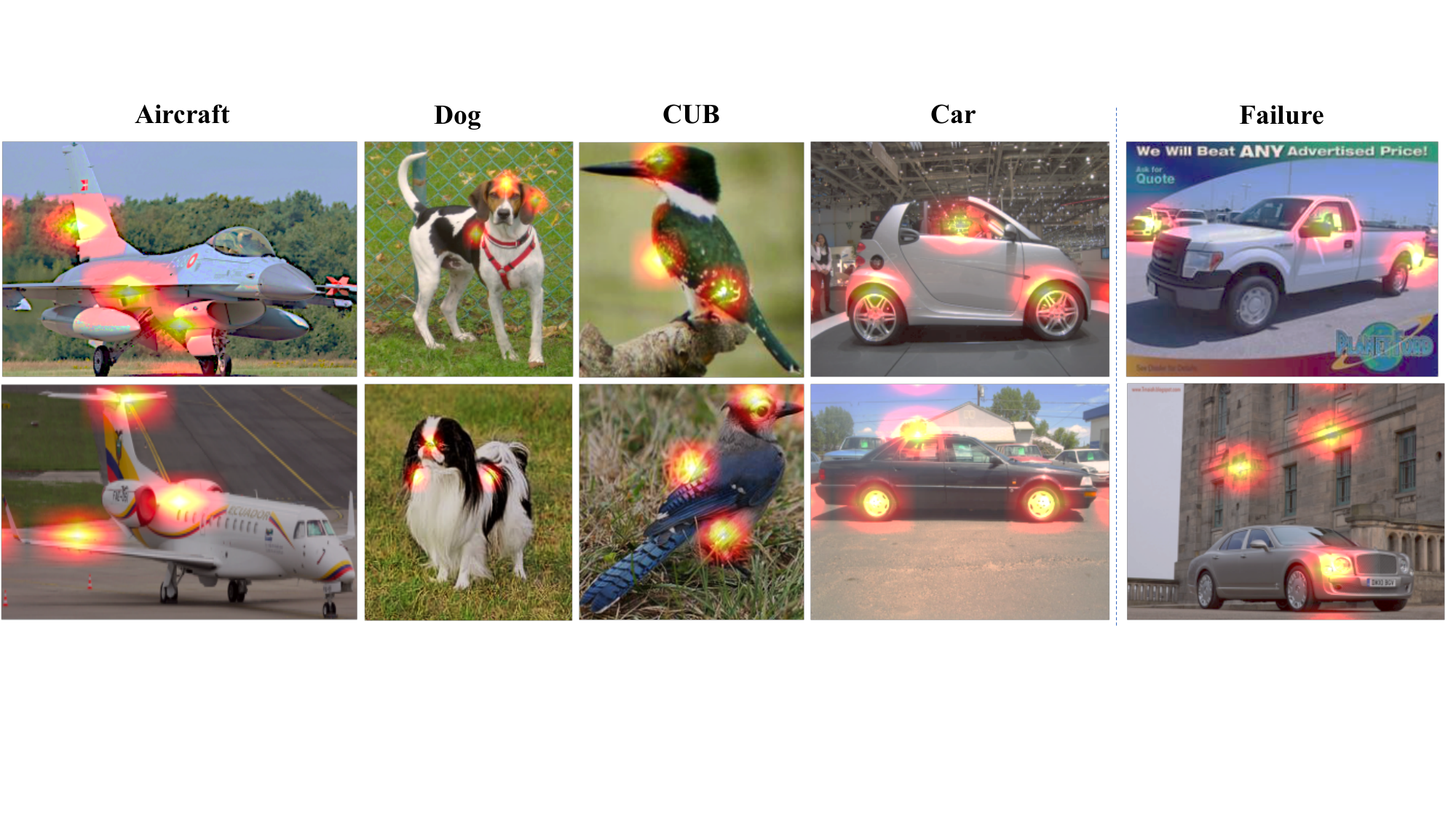}    \caption{Exemplar part locations learned by \ourmethod when $K=3$. From left to right: Aircraft, CUB, Dog, Car, and failure cases. \ourmethod{} can fail and locate parts on the background if it has visual signatures similar to an object.}
    \label{fig:parts}
\end{figure}

\noindent {\bf What parts does \ourmethod detect?} We visualize the locations $\mu_p$ learned by \ourmethod in \cref{fig:parts}. \ourmethod is able to detect consistent parts for the same task and often finds semantically meaningful parts like head and torso/breast in birds and dogs and wheels and doors/windows on cars. \cref{fig:parts} also shows some failure cases of \ourmethod{}, where it might fail to locate parts on the object if similar visual signatures appear in the background. A visualization of part templates learned and part expressions can be found in the supplementary (App C). 

\section{Conclusions}
\noindent We presented \ourmethod, a deep object-parsing method for fine-grained few-shot recognition. Our fundamental concept is that, while different object classes exhibit novel visual appearance, at a sufficiently small scale, visual patterns are duplicated. Hence, by leveraging training data to learn a dictionary of templates distributed across different relative locations, an object can be recognized simply by identifying which of the templates in the dictionary are expressed, and how these patterns are geometrically distributed. We build a statistical model for parsing that takes the output of a convolutional backbone as input to produce a parsed output. We then post-hoc learn to re-weight query and support instances to identify the best matching class, and as such this procedure allows for mitigating visual distortions. Our proposed method is an end-to-end deep neural network training method, and we show that our performance is not only competitive but also the outputs generated are interpretable.

\bibliography{iclr2023_conference}
\bibliographystyle{iclr2023_conference}

\appendix
\section*{Appendix}

\section{Derivation of \texttt{PARSE}}
As mentioned in Sec. 3 of the main paper, estimating part expression and location leads to two coupled optimization problems. 
\begin{equation} \label{supp.eq.part-exp}
z_{p}(\mu)=\arg\min_{\beta} \sum_{c \in C} \|\phi_{c, M(\mu)} - D_{p,c}\beta_{c}\|^2 + \lambda \|\beta\|_1.
\end{equation}
\begin{equation} \label{supp.eq.part-location}
\mu_p = \argmin_{\mu \in [G] \times [G]}\left [ L_p(\mu) \triangleq \sum_{c \in C} \|[\phi_c]_{M(\mu)} - D_{p,c}z_{p,c}(\mu)\|^2 + \lambda \|z_{p}(\mu)\|_1 \right ]
\end{equation}
For solving the above, we first approximate the solution to \cref{supp.eq.part-exp} by optimizing the reconstruction error and subsequently thresholding. As mentioned in the main paper, this is closely related to thresholding methods employed in LASSO~\cite{hastie01statisticallearning}. So, first we solve

\begin{align}
    z_p^\prime (\mu)=\arg\min_{\beta} \sum_{c \in C} \|\phi_{c, M(\mu)} - D_{p,c}\beta_{c}\|^2 \nonumber
\end{align}

As a reminder, the subscript $M(\mu)$ refers to the projection of $\phi_c$ onto the support of $M(\mu)$, which is an $s \times s$ grid centered at $\mu$. The quadratic form of the above optimization problem, gives us an explicit solution.

\begin{align} \label{supp.eq.z_prime}
    z_{p, c}^\prime (\mu) = \frac{(D_{p,c} \ast \delta_\mu) : \phi_c}{\|D_{p,c}\|^2} = \frac{(D_{p,c} \ast \phi_c)(\mu)}{\|D_{p,c}\|^2}
\end{align}

where $\delta_{\mu}(v) = \delta(\mu - v), v \in [G] \times [G]$ is a dirac delta centered at $\mu$, $\ast$ is a convolution\footnote{Note that following terminology from signal processing this is not actually a convolution but a cross-correlation. However, the way we use this term has been accepted in literature surrounding convolutional neural networks.} 
: $D_{p,c} \ast \delta_{\mu}(v) = \sum_{w} D_{p,c}(w-v) \delta_{\mu}(v)$ and `:' is the double-dot product or the sum of all elements of an element-wise/Hadamard product. 

For estimating location, we substitute $z_p^\prime$ into \cref{supp.eq.part-location} resulting in an upper bound for $L_p(\mu)$, which we denote as $L_p^\prime(\mu)$.

\begin{align} \label{supp.eq.L}
    L_p(\mu) \leq L_p^\prime(\mu) &= \sum_{c \in [C]}\|[\phi_{c,M(\mu)} - D_{p,c}z_{p,c}^\prime(\mu)\|^2 + \lambda \|z_{p}^\prime \|_1 \nonumber \\
    &= \sum_{c \in [C]} \left[ \|\phi_{c, M(\mu)}\|^2 - 2 (\phi_{c, M(\mu)} : D_{p,c}) z_{p,c}^\prime + \|D_{p,c}z_{p,c}^\prime\|^2 + \lambda |z_{p,c}^\prime| \right] \nonumber\\
    & \overset{(1)}{=} \sum_{c \in [C]} \left[ \|\phi_{c, M(\mu)}\|^2 - 2 (\phi_{c, M(\mu)} : D_{p,c}) \frac{(D_{p,c}\ast\phi_c)(\mu)}{\|D_{p,c}\|^2} \right. \nonumber \\
    & \qquad \qquad \left. + \|D_{p,c}\|^2 . \frac{(D_{p,c}\ast\phi_c)(\mu)^2}{\|D_{p,c}\|^4} + \lambda \frac{(D_{p,c}\ast\phi_c)(\mu)}{\|D_{p,c}\|^2} \right] \nonumber \\
    & \overset{(2)}{=} \sum_{c\in[C]} \left[ \|\phi_{c, M(\mu)}\|^2 - \frac{(D_{p,c}\ast\phi_c)(\mu)^2}{\|D_{p,c}\|^2} + \lambda \frac{(D_{p,c}\ast\phi_c)(\mu)}{\|D_{p,c}\|^2} \right] \nonumber \\
    &= \sum_{c\in[C]} \left[ \|\phi_{c, M(\mu)}\|^2 - \frac{(D_{p,c}\ast\phi_c)(\mu)^2}{\|D_{p,c}\|^2} + \lambda \frac{(D_{p,c}\ast\phi_c)(\mu)}{\|D_{p,c}\|^2} \right. \nonumber \\
    &\qquad\qquad\left.- \frac{\lambda^2}{4\|D_{p,c}\|^2} + \frac{\lambda^2}{4\|D_{p,c}\|^2} \right] 
\end{align}

For step (1) above, we substitute $z_{p,c}^\prime$ from \cref{supp.eq.z_prime}. For step (2), note that $D_{p,c} : \phi_{c, M(\mu)} = (D_{p,c} \ast \phi)(\mu)$, since $M(\mu)$ is an $s \times s$ attention map centered at $\mu$.

From \cref{supp.eq.L}, by ignoring the first and the last terms and contracting the binomial squares, we get the following as our estimate for $\mu_p$. Note that the last term is ignored because it does not depend on $\mu$. Also, the first term $\sum_{c\in[C]} \|\phi_{c, M(\mu)}\|^2$, which is the energy across all channels varies little for different values of $\mu$.
\begin{align} \label{supp.eq.mu_estimate0}
    \mu_p = \argmin_{\mu \in [G] \times [G]} L_{p}^\prime(\mu) &= \argmin_{\mu \in [G] \times [G]} -\sum_{c \in C} \left[ \frac{(D_{p,c}\ast\phi_c)(\mu)}{\|D_{p,c}\|} - \frac{\lambda}{\|D_{p,c}\|} \right]^2 \nonumber \\ &= \argmax_{\mu \in [G] \times [G]} \sum\limits_{c\in C}((\theta_{p,c}\ast\phi_c)(\mu)-\lambda_c)^2
\end{align}

$\theta_{p,c}=D_{p,c}/\|D_{p,c}\|$, and $\lambda_c = \lambda/2\|D_{p,c}\|$ becomes a channel dependent constant. The location estimate in \cref{supp.eq.mu_estimate0}, is thus, in the form of template matching per channel. $z_p$ is then estimated by substituting $\mu_p$ back in \cref{supp.eq.z_prime} to get $z_p^\prime$ and subsequently lower thresholding with $\zeta$.
\begin{align}\label{supp.eq.estimate_z0}
    z_{p,c}(\mu)=S_{\zeta}\left( \frac{(D_{p,c} \ast \phi_c)(\mu_p)}{\|D_{p,c}\|^2} \right)
\end{align}

\section{More on Compared Methods}
We compare \ourmethod to state-of-the-art few-shot learning methods, including RENet\cite{kang2021relational}, FRN\cite{wertheimer2021few},TDM\cite{lee2022task} and DeepEMD\cite{zhang2020deepemd} and also to methods like FOT\cite{wang2021fine}, VFD \cite{xu2021variational}, DN4\cite{li2019revisiting} and TDM\cite{lee2022task}, which are dedicated to the fine-grained setting. To highlight the contribution of \ourmethod, we tabulate in \cref{tab:method_compare} the differences of the model design compared to prior works \cite{tokmakov2019learning,hao2019collect,zhang2020deepemd,wu2021task} in few-shot learning that also use part composition. 

While there are prior works that learn recognition via object parts, and use instance-dependent reweighting, \ourmethod is unique in using reconstruction with templates (RwT) as a criterion, uses a prior on the geometry of parts using part-locations and uses this geometry for comparing instances. See \cref{tab:method_compare} for a tabulated comparison.

\begin{table}[h]
    \caption{Components compared to prior works. recognition using parts; RwT: Reconstruction with Templates; Geo: using geometry of parts for instance comparison, and incorporating prior on geometry.; Reweighting: instance dependent reweighting of matching scores. Prior part-based FSL methods do not employ some components proposed in \ourmethod.}
    \vspace{1mm}
    \centering
    \setlength{\tabcolsep}{2mm}
    \begin{tabular}{l c c  c c}
        \toprule
        Methods & Parts & RwT & Geo & Reweighting \\
        \midrule
        LCR\cite{tokmakov2019learning} & & & &\\
        SAML\cite{hao2019collect} & \checkmark & & & \checkmark \\
        DeepEMD\cite{zhang2020deepemd}  & & & & \checkmark \\
        TPMS\cite{wu2021task} & \checkmark & & & \checkmark \\
        TDM\cite{lee2022task} & & & & \checkmark \\
        \ourmethod (ours) & \checkmark & \checkmark & \checkmark & \checkmark \\
        \bottomrule
    \end{tabular}
    \label{tab:method_compare}
\end{table}

\section{Visualizing Templates and Part Expressions}
\begin{figure}[h]
    \centering
    \includegraphics[trim=0cm 0cm 0cm 0cm,clip,width=0.8\linewidth]{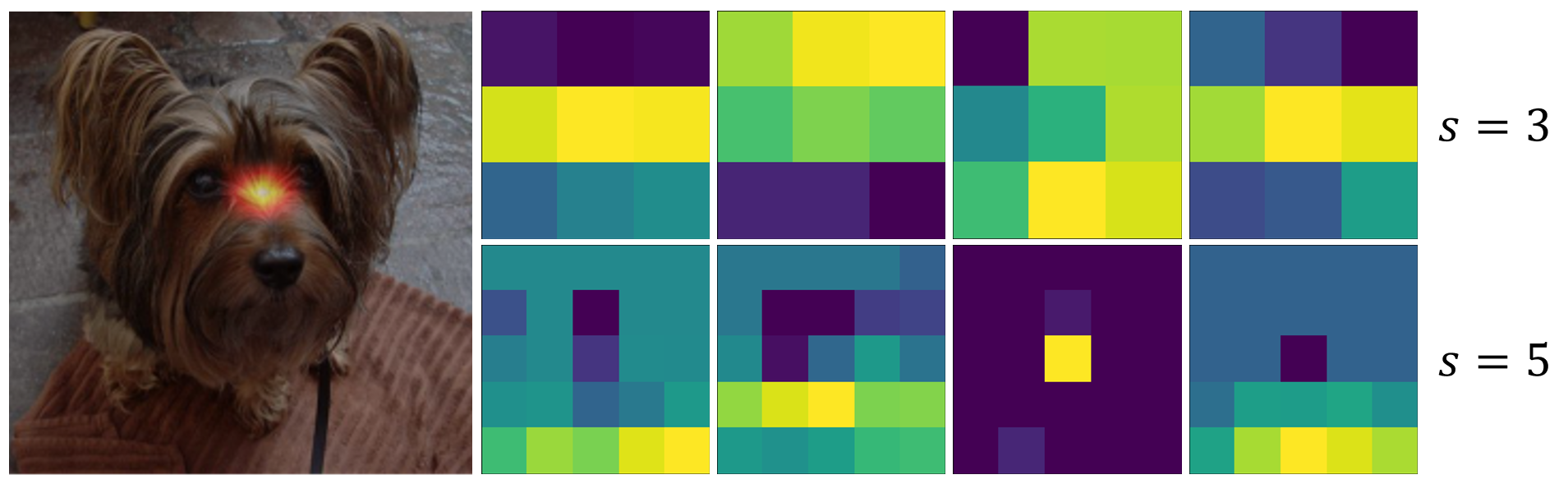}
    \caption{Exemplar templates of learned dictionary $D_p$. The templates shown are for randomly sampled channels for scale $3$ (top) and $5$ (bottom).}
    \label{fig:texture}
\end{figure}

\begin{figure}[h]
    \centering
    \includegraphics[trim=0cm 2.1cm 0cm 2.5cm,clip,width=1\linewidth]{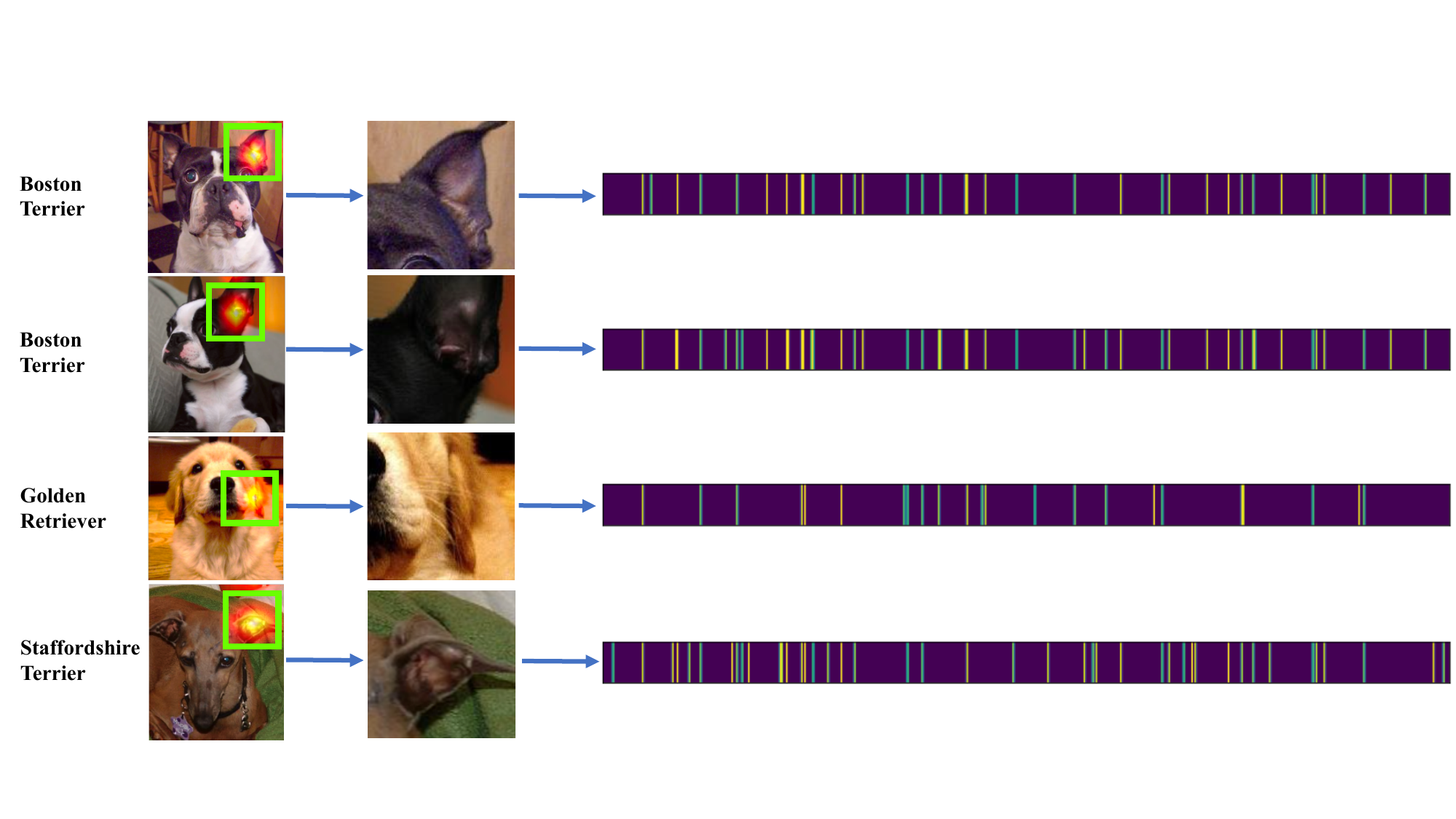}
    \caption{Template coefficients $z_{p}$ of the same part for two Boston Terriers (top 2 rows), a Golden Retriever (3rd row) and a Staffordshire Terrier (4th row). Template coefficients for images of the same class are similar. Visually-similar classes (Boston Terrier and Staffordshire Terrier) share some of the same activated templates, while visually distinct classes (Golden Retriever) differ a lot on their selection of active templates.}
    \label{fig:dict}
\end{figure}

Some templates of the learned dictionary $D_p$ are visualized in \cref{fig:texture}. Our model uses each template to reconstruct the original feature in the corresponding channel. We see diverse visual representations in different channels, implying that \ourmethod learns diverse visual templates from the training set to express objects. \cref{fig:dict} shows the activated templates for different objects. The model uses the same templates to express the same class.

\end{document}